\documentclass[runningheads]{llncs}

\usepackage{eccv}

\usepackage{eccvabbrv}

\usepackage{graphicx}
\usepackage{booktabs}
\usepackage{multirow}
\usepackage{placeins}
\usepackage[breakable]{tcolorbox}

\newtcolorbox{promptbox}[2][blue]{
  colback=#1!5,
  colframe=#1!60,
  fonttitle=\bfseries,
  title={#2},
  boxrule=0.5pt,
  arc=2pt,
  left=6pt, right=6pt, top=4pt, bottom=4pt,
  fontupper=\small\ttfamily,
  breakable,
}

\usepackage[accsupp]{axessibility}

\usepackage[pagebackref,breaklinks,colorlinks,citecolor=eccvblue]{hyperref}

\usepackage{orcidlink}

\begin{document}

\title{FED-Bench: A Cross-Granular Benchmark \\ for Disentangled Evaluation of \\ Facial Expression Editing} 

\titlerunning{FED-Bench}

\author{Fengjian Xue\inst{1}\thanks{\ Equal contribution.} \and
Xuecheng Wu\inst{1}$^*$ \and
Heli Sun\inst{1}\thanks{\ Corresponding author.} \and
Yunyun Shi\inst{1} \and \\
Shi Chen\inst{1} \and
Liangyu Fu\inst{2} \and
Jinheng Xie\inst{3}  \and
Dingkang Yang\inst{4}  \and \\
Hao Wang\inst{1}$^{**}$ \and
Junxiao Xue\inst{5}  \and
Liang He\inst{1}
}

\authorrunning{F. Xue et al.}

\institute{Xi'an Jiaotong University \and
Northwestern Polytechnical University \and
National University of Singapore \and
Fudan University \and
Zhejiang Lab \\
\email{\{hlsun, haowangx, lhe\}@xjtu.edu.cn}
}

\maketitle

\begin{abstract}
Facial expression image editing requires fine-grained control to strictly preserve human identity and background while precisely manipulating expression. However, existing editing benchmarks primarily focus on general scenarios, lacking high-quality facial images and corresponding editing instructions. Furthermore, current evaluation metrics exhibit systemic biases in this task, often favoring lazy editing or overfit editing. To bridge these gaps, we propose FED-Bench, a comprehensive benchmark featuring rigorous testing and an accurate evaluation suite. First, we carefully construct a benchmark of 747 triplets through a cascaded and scalable pipeline, each comprising an original image, an editing instruction, and a ground-truth image for precise evaluation. Second, we introduce FED-Score, a cross-granularity evaluation protocol that disentangles assessment into three dimensions: Alignment for verifying instruction following, Fidelity for testing image quality and identity preservation, and Relative Expression Gain for quantifying the magnitude of expression changes, effectively mitigating the aforementioned evaluation biases. Third, we benchmark 18 image editing models, revealing that current approaches struggle to simultaneously achieve high fidelity and accurate expression manipulation, with fine-grained instruction following identified as the primary bottleneck. Finally, leveraging the scalable characteristic of introduced benchmark engine, we provide a 20k+ in-the-wild facial training set and demonstrate its effectiveness by fine-tuning a baseline model that achieves significant performance gains. Our benchmark and related code will be made publicly open soon.
\keywords{Benchmark \and Baseline \and Facial expression \and Image editing}
\end{abstract}

\begin{figure}[t!]
  \centering
  \includegraphics[width=\textwidth]{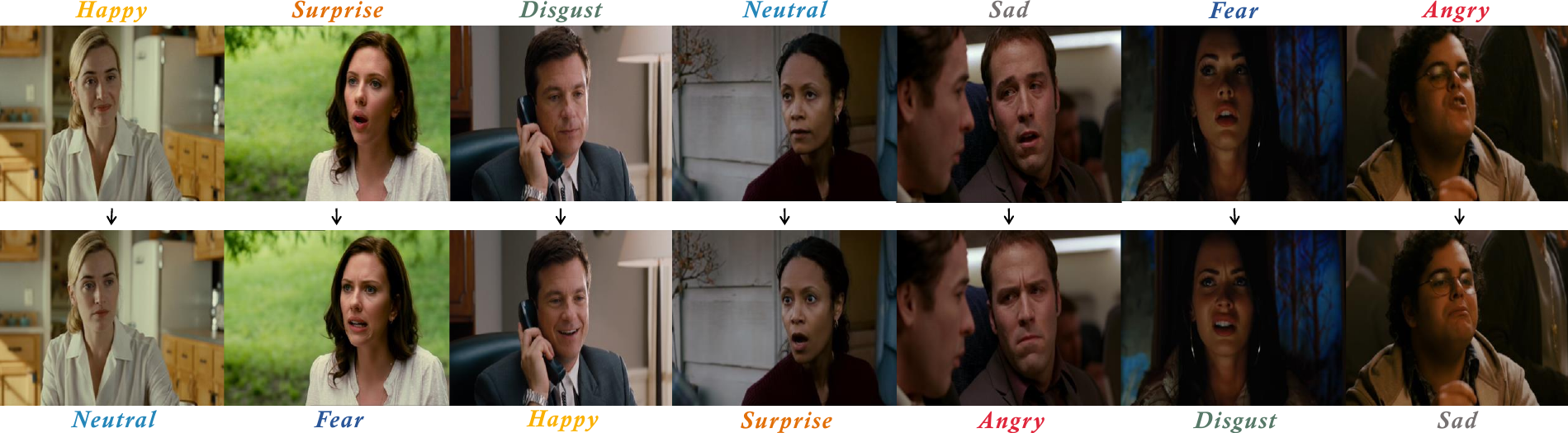}
  \caption{The overview of FED-Bench. The figure illustrates representative samples from our benchmark covering seven basic emotions.}
  \vspace{-1.5em}
  \label{fig:overview}
\end{figure}

\section{Introduction}
\label{sec:intro}

Recently, driven by the rapid advancement of generative technologies such as Diffusion Models\cite{ho2020denoising,textpecker,han2025beyond} and Multimodal Large Language Models (MLLMs)~\cite{xue2026towards,wang2025jtd,wu2025vic}, text-guided image editing has achieved remarkable success. Users can now perform rich modifications to image content by simply providing straightforward text prompts. Among its various applications, Facial Expression Image Editing has drawn significant attention due to its immense potential in areas like real-time interaction and visual effects for film and television. However, unlike general image editing tasks such as altering object colors or performing global style transfers, facial expression editing is highly challenging. It requires models to strictly preserve the subject's identity and background information while achieving fine-grained control over expressions. Due to the fine-grained nature of this task, most existing general-purpose editing models struggle to balance these requirements, often resulting in distorted facial features or unintended modifications to non-target regions.

Although facial expression editing technology is evolving rapidly, further development in this field is severely hindered by the lack of rigorous evaluation benchmarks~\cite{wu2025avf,wu2025towards,yang2025improving}. On the one hand, prevailing image editing benchmarks (\textit{e.g.}, Imgedit\cite{ye2025imgedit}, EditBench\cite{wang2023imagen}, and I2ebench\cite{ma2024i2ebench}) primarily focus on general scenarios. They severely lack high-quality facial samples and do not design corresponding editing instructions for the subtle movements of facial muscles. On the other hand, and more critically, most current benchmarks only provide a source image and a text instruction, lacking a Ground Truth (GT) reference image for rigorous comparison. Without pixel-level GT as an anchor, existing evaluations remain at a coarse-grained or subjective level, making it impossible to conduct objective and quantitative assessments of a model's true performance in fine-grained expression editing tasks.

In addition to the absence of benchmark data, currently widely-used image quality evaluation metrics exhibit severe systemic biases when applied to facial expression editing tasks~\cite{wu2025scalable}. Specifically, the existing evaluation systems often face a dilemma: if heavily weighted towards image fidelity metrics (\textit{e.g.}, DINO Score\cite{caron2021emerging}), the evaluation protocol favors Lazy Editing, where a model receives an exceptionally high score by outputting the source image directly without any modification. Conversely, if heavily weighted towards semantic alignment metrics (\textit{e.g.}, CLIP Score\cite{hessel-etal-2021-clipscore}), the evaluation system easily falls into Overfit Editing, where the model generates overly exaggerated and distorted expressions to cater to the text prompt, even at the expense of the subject's original identity. Such persistent evaluation biases not only fail to reflect the true editing capabilities of the models but may also mislead the design direction of future models.

To break this dilemma and compensate for the deficiencies in existing benchmarks and evaluation systems, we propose FED-Bench: a comprehensive benchmark framework specifically tailored for fine-grained facial expression editing. We collect and clean 747 high-quality data triplets. Each triplet consists of a source image, an editing instruction, and an accurately paired Ground Truth image. Addressing the specific nature of this task, we design a rigorous data filtering pipeline to obtain the GT images, thereby providing a reliable reference for objective and accurate performance evaluation. 

Furthermore, based on introduced FED-Bench, we innovatively introduce FED-Score, a novel cross-granular evaluation protocol. Discarding the ambiguity of single metrics, FED-Score explicitly decouples the evaluation process into three independent yet complementary dimensions: Alignment for verifying instruction execution, Fidelity for ensuring overall image quality and identity preservation, and Relative Expression Gain for precisely quantifying the magnitude of expression changes. Through the mutual constraint and joint assessment of these three dimensions, our protocol establishes a reliable and robust evaluation framework that significantly mitigates evaluation biases such as lazy editing and overfit editing.

Extensive experiments demonstrate that compared to existing image evaluation metrics, FED-Score exhibits higher consistency with human subjective judgment. Meanwhile, benchmarking based on FED-Bench profoundly reveals the performance bottlenecks of current mainstream image editing models regarding fine-grained control. To further facilitate continuous research in this field, we leverage the scalability of our automated pipeline to construct a large-scale training set of 20k+ in-the-wild facial expression editing pairs. Fine-tuning an existing model on this data yields substantial improvements in both expression accuracy and fidelity, demonstrating the practical value of the proposed data construction paradigm.

In summary, the main contributions of this paper are as follows:
\begin{enumerate}
    \item \textbf{Construction of FED-Bench:} We propose the first fine-grained test benchmark specifically designed for facial expression editing. It contains 747 high-quality triplets (source image, editing instruction, Ground Truth), providing the necessary data foundation for precise evaluation.
    \item \textbf{Proposal of the FED-Score Evaluation Protocol:} We design a cross-granular, decoupled evaluation system that comprehensively assesses alignment, fidelity, and relative expression gain. This effectively overcomes the systemic biases (\textit{e.g.}, favoring lazy editing and overfit editing) prevalent in existing metrics.
    \item \textbf{In-depth Evaluation and Analysis:} We conduct comprehensive evaluations of state-of-the-art image editing models, revealing their performance bottlenecks in balancing identity preservation with precise editing.
    \item \textbf{Open-sourcing a Large-scale Training Set:} We provide a high-quality, in-the-wild facial image training set containing 20k+ samples to facilitate the development of finer and more robust facial editing models in the future.
\end{enumerate}

\section{Related Work}
\label{sec:related}

\subsection{Image Editing and Facial Expression Manipulation}
General image editing has undergone a profound evolution from early GAN/VAE-based attribute manipulation to instruction-driven editing powered by large-scale diffusion models and autoregressive backbones. InstructPix2Pix\cite{brooks2023instructpix2pix} established the mainstream framework of training conditional diffusion models on synthetic paired data. AnyEdit\cite{yu2025anyedit} introduced task-aware routing and mixture-of-experts modules to support more diverse editing requirements. OmniGen\cite{xiao2025omnigen} adopted a single Transformer backbone for joint text-image encoding, eliminating the need for external encoders entirely. On the multimodal language-vision collaboration front, MGIE\cite{fu2023guiding} leverage MLLMs to provide precise instruction grounding signals. Step1X-Edit\cite{liu2025step1x} significantly improves complex instruction following and background consistency preservation through a dual-stream bridging mechanism. ReasonEdit\cite{yin2025reasonedit} and related works further incorporate reasoning loops into the editing pipeline. Overall, the above methods are continuously advancing toward open-ended, text-guided image editing with stronger semantic controllability and fidelity.

Facial expression manipulation presents greater challenges than general scene editing, as it demands high-precision control over local geometric structures and subtle appearance cues while strictly preserving identity and background consistency. Early text-driven methods such as StyleCLIP\cite{patashnik2021styleclip} demonstrated promising qualitative results, but tend to cause unintended identity alterations when the magnitude of instruction-driven changes is large. To address this, subsequent works introduced structured geometric conditions: MagicFace\cite{wang2024magicface} conditions on action unit (AU) deltas paired with a dedicated identity encoder to achieve fine-grained expression editing; LaTo\cite{zhang2026latolandmarktokenizeddiffusiontransformer} encodes facial landmarks as semantic tokens to enable precise pose and expression control while preserving identity; DiffFERV\cite{chen2025diffferv} specializes diffusion priors to the face domain and incorporates temporal modeling to achieve high-fidelity expression editing on video sequences.

Despite achieving promising results in specific settings, these methods still rely primarily on supervised labels, action unit annotations, or geometric priors, and are not designed end-to-end for natural language expression instructions, making it difficult to capture the nuanced emotional descriptions present in various user prompts. 

\subsection{Image Editing Datasets and Benchmarks}

The evaluation of instruction-based image editing has rapidly evolved to address the growing complexity of user intents and the demand for higher visual fidelity. To capture realistic human instructions, datasets like MagicBrush\cite{zhang2023magicbrush} provide manually annotated instruction-image pairs for real-world scenarios, ensuring high alignment with human intent. Meanwhile, HQ-Edit\cite{hui2024hq} focuses on high-resolution data collection to guarantee the visual quality of the edited outcomes. To encompass a wider range of editing semantics, AnyEdit\cite{yu2025anyedit} constructs a comprehensive dataset that integrates various operations, such as object addition, removal, and replacement, into a unified framework.

Beyond dataset construction, recent evaluation paradigms have shifted towards assessing holistic model capabilities on harder, real-image distributions. Modern benchmarks, such as RISEBench\cite{zhao2025envisioning} and SmartEdit\cite{huang2024smartedit}, have been proposed to systematically evaluate models on complex reasoning, spatial understanding, and multi-dimensional instruction following. Together, these protocols provide standardized comparisons for multi-turn edits and compositional instructions, reflecting the field's growing need for robust evaluation in open-ended editing tasks.

\subsection{Evaluation Metrics for Image Editing}

Evaluating AI-generated and edited images requires assessing both visual fidelity and text-image alignment. Traditional measures like the Inception Score (IS)\cite{10.5555/3157096.3157346} and Fréchet Inception Distance (FID)\cite{10.5555/3295222.3295408} are widely employed to evaluate overall realism and image quality. For measuring the semantic alignment between the edited image and the instruction or text prompt, metrics such as the CLIP Score\cite{hessel-etal-2021-clipscore} and BLIP Score\cite{10.5555/3618408.3619222} have become the established standards to ensure the generated content accurately reflects the textual guidance.

Unlike standard text-to-image generation, subject-driven image editing, especially facial expression manipulation, strictly demands the preservation of the source image's structural integrity and the subject's identity. To measure this perceptual and structural consistency, Learned Perceptual Image Patch Similarity (LPIPS)\cite{zhang2018unreasonable} is frequently used, while deep self-supervised feature extractors like DINO\cite{caron2021emerging} and specialized face recognition models\cite{deng2019arcface} evaluate semantic consistency and robust identity preservation. Recently, Large Multimodal Models (LMMs) such as GPT-4o have also emerged as powerful automatic evaluators\cite{ku2024viescore}, leveraging their advanced reasoning and instruction-following capabilities to comprehensively assess text-image alignment and evaluate the nuanced visual changes inherent in complex facial manipulation tasks.
\section{FED-Bench Construction}
\label{sec:bench}

\begin{figure}[t!]
    \centering
    \includegraphics[width=\textwidth]{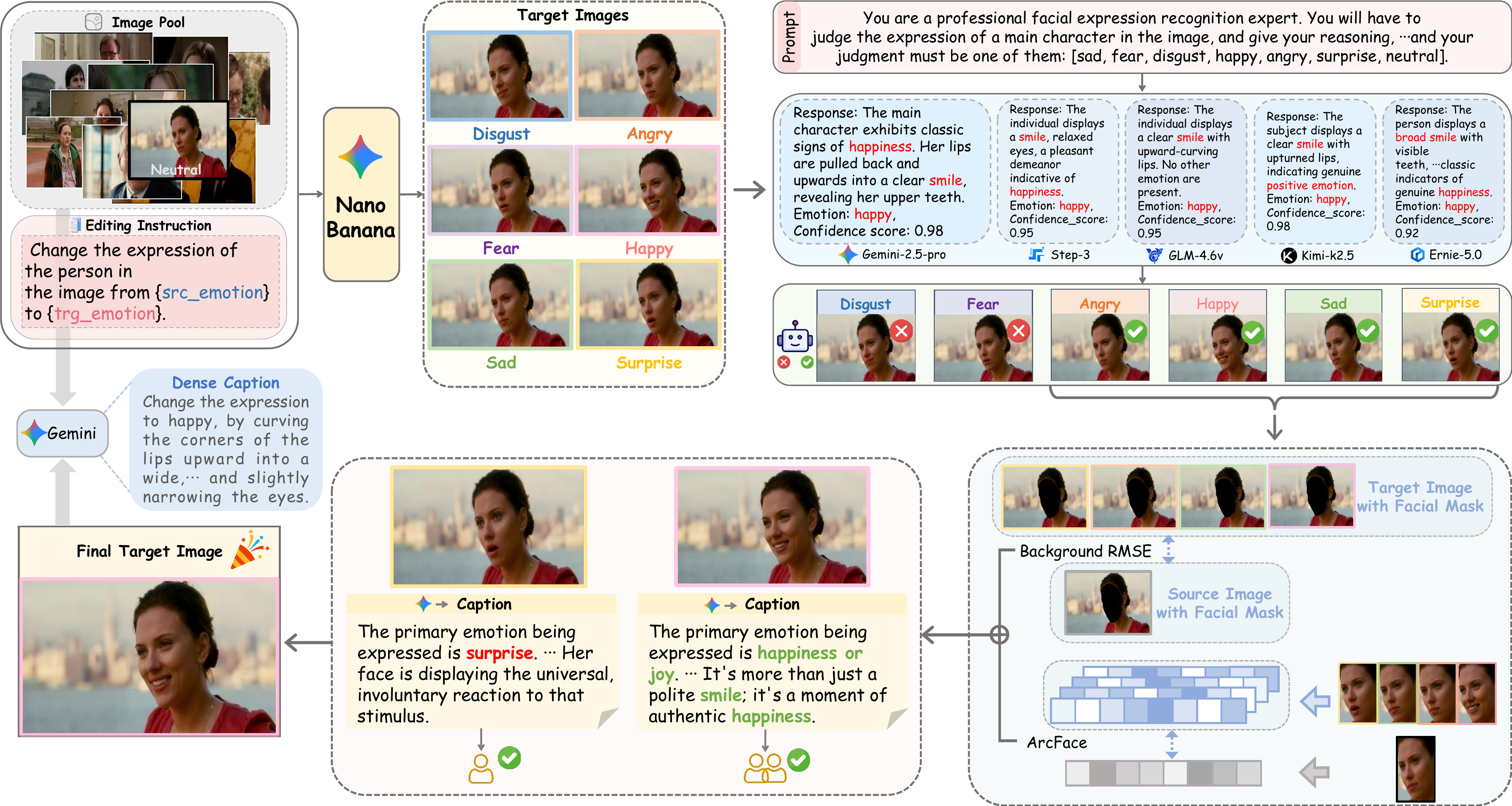}
    \caption{The overall illustrations of FED-Bench construction pipeline.}
    \label{fig:pipeline}
\end{figure}

To obtain high-quality Ground Truth images for refined evaluation, we design a rigorous multi-stage screening pipeline. As shown in \cref{fig:pipeline}, the pipeline consists of five stages: source data screening, candidate target image generation, coarse-grained expression recognition, fidelity ranking, and human verification.

\noindent\textbf{Source Data.} Facial expression editing demands source images with rich in-the-wild backgrounds and high-quality facial details, so that both background preservation and identity preservation can be rigorously tested. We select SFEW 2.0\cite{dhall2011static} and DFEW\cite{jiang2020dfew} as raw data sources, as their movie-clip origins naturally provide diverse backgrounds and sufficient clarity. After filtering out images with wrong expression label, low resolution, severe occlusion, or extreme lighting, we curate 747 high-quality source images for FED-Bench.

\noindent\textbf{Candidate target image generation.} Due to the scarcity of perfectly paired in-the-wild facial expression images, we use a state-of-the-art editing model $\mathcal{G}$ (Nano Banana) to generate candidate targets. Our benchmark covers seven basic expressions: angry, disgust, fear, happy, neutral, sad, and surprise. For each source image $I_{\text{src}}$, we generate candidates for the remaining six expressions using the template ``change the expression from \{src\_emotion\} to \{trg\_emotion\}'':
\begin{equation}
    I_{\text{cand}}^{(i)} = \mathcal{G}(I_{\text{src}}, T(src\_emotion, trg\_emotion(i))), \quad i = 1, 2, \ldots, 6.
\end{equation}

\noindent\textbf{Coarse-grained facial expression recognition.} 
We first verify whether candidates successfully manifest the target expressions. Expert FER models generalize poorly in the wild ($<$40\% accuracy, \cref{tab:fer_comparison}), so we adopt MLLMs instead. However, even the best MLLM (Gemini-2.5-Pro) achieves only 64.93\% on fine-grained 7-class classification, and strict filtering at this accuracy would discard many valid samples.
To address this, we propose a \textbf{Coarse-grained Expression Recognition Strategy} that regroups the seven expressions into three polarities: Positive (happy), Neutral, and Negative (remaining five), boosting peak accuracy to 81.23\% (\cref{tab:fer_comparison}). We further introduce a \textbf{Multi-model Voting Mechanism} using a 5-model ensemble, which achieves optimal performance (\cref{tab:voting_fer}). This combined strategy effectively eliminates invalid samples that fail to exhibit the desired expression changes. Comprehensive test results are presented in supplementary materials.

\begin{table}[t]
    \centering
    \begin{minipage}[t]{0.38\textwidth}
        \centering
        \caption{Comparison of facial expression recognition accuracy.}
        \label{tab:fer_comparison}
        \resizebox{\linewidth}{!}{%
        \setlength{\tabcolsep}{6pt}
        \begin{tabular}{lcc}
            \toprule
            Model & Fine & Coarse \\
            \midrule
            \multicolumn{3}{l}{\textit{Expert Models}} \\
            BEiT\cite{tanneru2025beit_affectnet, bao2021beit}        & 35.7\% & -- \\
            Luxand      & 39.8\% & -- \\
            DeepFace\cite{serengil2026boosted}    & 36.5\% & -- \\
            \midrule
            \multicolumn{3}{l}{\textit{MLLMs}} \\
            Gemini-2.5-Pro       & \textbf{64.93\%} & \underline{81.23\%} \\
            Doubao-Seed-1.6-Vision & \underline{62.52\%} & 80.59\% \\
            ERNIE-5.0-Thinking   & 59.17\% & 80.76\% \\
            GLM-4.6V             & 58.77\% & 77.48\% \\
            Kimi-K2.5            & 57.83\% & \textbf{81.79\%} \\
            Step-3               & 57.56\% & 79.76\% \\
            Claude-Sonnet-4.5    & 53.55\% & 76.31\% \\
            Qwen3-VL-Plus        & 48.59\% & 76.68\% \\
            Claude-Haiku-4.5     & 48.19\% & 71.75\% \\
            Grok-4.1             & 43.51\% & 66.53\% \\
            \bottomrule
        \end{tabular}%
        }
    \end{minipage}
    \hfill
    \begin{minipage}[t]{0.58\textwidth}
        \centering
        \caption{Accuracy comparison of different voting ensemble sizes. The optimal 5-model in Fine-grained FER ensemble consists of Gemini-2.5-Pro, GLM-4.6V, Kimi-K2.5, ERNIE-5.0, and Step-3. The optimal in Coarse-grained FER ensemble consists of Gemini-2.5-Pro, GLM-4.6V, Kimi-K2.5, ERNIE-5.0, and Doubao-seed-1.6-vision.}
        \label{tab:voting_fer}
        \vspace{4pt}
        \resizebox{\linewidth}{!}{%
        \setlength{\tabcolsep}{6pt}
        \begin{tabular}{l cc cc}
        \toprule
        \multirow{2}{*}{Setting} & \multicolumn{2}{c}{Fine-grained (7-cls)} & \multicolumn{2}{c}{Coarse-grained (3-cls)} \\
        \cmidrule(lr){2-3} \cmidrule(lr){4-5}
         & Best Acc & Avg Acc & Best Acc & Avg Acc \\
        \midrule
        3-model      & 66.13 & 59.94 & 84.47 & 81.04 \\
        \textbf{5-model}      & \textbf{66.67} & 61.75 & \textbf{84.87} & 82.27 \\
        7-model      & 66.00 & 62.78 & 84.61 & 82.88 \\
        9-model      & 63.99 & 63.25 & 83.40 & 82.90 \\
        \bottomrule
        \end{tabular}%
        }
    \end{minipage}
\end{table}

\noindent\textbf{Fidelity Ranking.} 
In this stage, our evaluation encompasses two complementary dimensions. First, to measure identity preservation (ID), we utilize the pre-trained face recognition network ArcFace\cite{deng2019arcface} to extract identity feature vectors from both the source and candidate images, subsequently calculating their cosine similarity $S_{\text{id}}^{(i)}$. Second, to assess pixel-level background consistency, we compute the Root Mean Squared Error (RMSE) across all pixels outside the facial expression region, denoted as $S_{\text{bg}}^{(i)}$. 
After normalizing both distance metrics, we combine them into an overall preservation score through a weighted fusion:
\begin{equation}
    S_{\text{total}}^{(i)} = \text{Norm}(S_{\text{id}}^{(i)}) + \text{Norm}(S_{\text{bg}}^{(i)}).
\end{equation}
Based on this score, we sort all expression-verified samples in descending order and strictly retain only the top two candidate images as potential Ground Truth references for the final stage.

\noindent\textbf{Human Verification.} 
Automated filtering cannot capture subtle facial nuances and microscopic artifacts, so we incorporate a human-in-the-loop verification as the final quality assurance step. We first use an MLLM to generate a detailed expression caption for each candidate image as an objective reference. Three human evaluators then independently inspect the candidate pairs, considering expression naturalness, instruction alignment, and identity/background preservation, with the final Ground Truth determined by majority voting.

\noindent\textbf{Dense Instruction Generation.}
We introduce a mechanism to generate fine-grained editing instructions. This serves two primary purposes: to further refine the control dimensions of facial expression editing, and to comprehensively evaluate the instruction-following capabilities of image editing models when presented with detailed textual guidance.
Specifically, we employ MLLM as a visual difference analyzer. We input the paired source image and its corresponding GT image into the MLLM, prompting it to meticulously compare the specific facial expression alterations between the two images (e.g., the movement of facial muscles, the curvature of the eyes, and the opening or closing of the mouth). Based on this cross-image difference analysis, the MLLM generates a highly descriptive and fine-grained editing instruction for each data triplet. Compared to the simple instruction used in earlier stages, these dense instructions provide more precise guidance for generative models and offer a more challenging linguistic input for subsequent performance evaluation.

\noindent\textbf{Scalability and Training Set.}
Our automated pipeline also exhibits excellent scalability. We further selected 20k+ in-the-wild facial images from RAF\cite{li2019reliable} and DFEW, using Qwen-Image-Edit-2511 as the generator, to efficiently construct a large-scale training set. Fine-tuning Flux-Kontext-dev on this data yields significant improvements in both expression precision and fidelity (\cref{subsec:benchtest}), validating the practical value of our data construction paradigm.

\section{The FED-Score Evaluation Suite}
\label{sec:evaluation}

To comprehensively and fairly evaluate facial expression editing models, it is imperative to break away from the traditional reliance on single-dimensional metrics such as standalone CLIP-Scores\cite{hessel-etal-2021-clipscore} or LPIPS\cite{zhang2018unreasonable}. Sole reliance on these isolated measurements often allows models to exploit evaluation loopholes, resulting in either lazy editing where the model fails to make necessary changes, or overfit editing where it generates exaggerated distortions. 
To address this, we propose the FED-Score, a decoupled evaluation protocol that deeply integrates rule-based (ID, BG, REG) computations with the model-based (PQ, SC, GTA) perceptual capabilities of MLLMs. Inspired by VIEScore\cite{ku2024viescore}, we establish a robust MLLM-based scoring framework for our perceptual metrics. The details of prompt templates are available in supplementary material.
Specifically, we explicitly decompose the comprehensive evaluation into three distinct dimensions: Fidelity, Alignment, and Relative Expression Gain.

\subsection{Fidelity Metrics}
\label{subsec:metric_fidelity}

The Fidelity dimension is designed to assess whether the model faithfully preserves elements that should remain unchanged while maintaining high overall visual quality. Our fidelity score ($\mathcal{S}_{Fid}$) is formulated by combining three distinct components: Identity Preservation (ID), Background Consistency (BG), and Perceptual Quality (PQ).

\noindent\textbf{Identity Preservation (ID).} The fundamental premise of facial editing is that the subject's identity must be preserved. We extract identity feature vectors from the source image $I_{\text{src}}$ and the target image $I_{\text{trg}}$ using a pre-trained ArcFace network, denoted as $\text{ArcFace}(\cdot)$. We then calculate their cosine similarity to serve as the ID score, which can be defined as:
\begin{equation}
    \text{ID}(I_{\text{src}}, I_{\text{trg}}) = \cos\!\left(\text{ArcFace}(I_{\text{src}}),\; \text{ArcFace}(I_{\text{trg}})\right).
\end{equation}

\noindent\textbf{Background Consistency (BG).} Since facial expression modifications should be strictly confined to the facial region, we first calculate the Root Mean Square Error (RMSE) of pixels exclusively in the non-facial areas, defined by a background mask $M_{\text{bg}}$, between $I_{\text{src}}$ and $I_{\text{trg}}$. To align with the evaluation logic where a higher score indicates better performance, we formulate the BG score by applying a normalization function $\text{Norm}(\cdot)$, \textit{i.e.},
\begin{equation}
    \text{BG}(I_{\text{src}}, I_{\text{trg}}) = \text{Norm}\!\left( \sqrt{\frac{1}{|M_{\text{bg}}|} \sum_{(x,y) \in M_{\text{bg}}} \left(I_{\text{src}}(x,y) - I_{\text{trg}}(x,y)\right)^2} \right),
\end{equation}
the detailed implementation of the $\text{Norm}(\cdot)$ function is provided in the supplementary material.

\noindent\textbf{Perceptual Quality (PQ).} Beyond physical pixel preservation, we employ an MLLM as an advanced visual evaluator to score the overall perceptual quality of $I_{\text{trg}}$, checking for unnatural artifacts or degradation. 

By integrating these three complementary metrics, we formulate the comprehensive fidelity score $\mathcal{S}_{\text{fid}}$ as their arithmetic mean:
\begin{equation}
    \mathcal{S}_{\text{fid}} = \text{Mean} \Big( \text{ID}(I_{\text{src}}, I_{\text{trg}}),\, \text{BG}(I_{\text{src}}, I_{\text{trg}}),\, \text{PQ}(I_{\text{trg}}) \Big).
\end{equation}

\subsection{Alignment Metrics}
\label{subsec:metric_alignment}
The Alignment dimension evaluates how accurately the generated image executes the editing instructions and manifests the desired expression. It comprises two core components: Semantic Consistency (SC) and GT-based Expression Alignment (GTA).

\noindent\textbf{Semantic Consistency (SC).} We prompt the MLLM to act as a rigorous judge to evaluate the semantic matching degree between the generated target image $I_{\text{trg}}$ and the input text instruction $T$.

\noindent\textbf{GT-based Expression Alignment (GTA).} Benefiting from the precise reference images provided by our FED-Bench, we transcend text-only evaluation. We leverage the MLLM to directly compare $I_{\text{trg}}$ with the reference image $I_{\text{gt}}$, specifically scoring the similarity of their facial expressions. This image-to-image semantic supervision significantly enhances the reliability of the alignment assessment.

Similarly, we integrate these two dimensions to formulate the comprehensive alignment score $\mathcal{S}_{\text{align}}$, defined as their arithmetic mean:
\begin{equation}
    \mathcal{S}_{\text{align}} = \text{Mean} \Big( \text{SC}(I_{\text{trg}}, T),\, \text{GTA}(I_{\text{trg}}, I_{\text{gt}}) \Big).
\end{equation}

\subsection{Relative Expression Gain}
\label{subsec:metric_reg}

The primary vulnerability of existing image editing evaluation protocol is their susceptibility to lazy editing. If a model simply outputs the source image without any modification, it achieves perfect fidelity scores (e.g., ID and BG metrics). To fundamentally resolve this systemic flaw, we introduce the Relative Expression Gain (REG), a novel quantitative metric designed to measure the actual magnitude of expression change relative to the expected change provided by our benchmark.

We first compute the perceptual distance of the facial region between the generated target image and the source image using LPIPS\cite{zhang2018unreasonable}, which represents the model's actual expression alteration. We then normalize this distance by the optimal perceptual change derived from our precise Ground Truth. To penalize both lazy editing and overfit editing, we formulate the final REG score $\mathcal{S}_{\text{reg}}$ as a Gaussian-like penalty function centered at 1.0:
\begin{equation}
    \text{REG} = \frac{\text{LPIPS}_{\text{face}}(I_\text{src},\; I_\text{trg})}{\text{LPIPS}_{\text{face}}(I_\text{src},\; I_\text{gt})}, \quad
    \mathcal{S}_{\text{reg}} = \exp\!\left(-\frac{(\text{REG} - 1)^2}{2\sigma^2}\right),
\end{equation}
where $\sigma$ is a hyperparameter controlling the tolerance width of the penalty, which is empirically set to $0.5$ in our evaluation. In this formulation, the model achieves the maximum score of 1.0 \textit{only} when its expression alteration magnitude perfectly matches that of the Ground Truth. Consequently, this metric naturally penalizes both lazy editing and overfit editing.

\subsection{The Comprehensive FED-Score}
\label{subsec:comprehensive_fedscore}

Having established three orthogonal evaluation dimensions, we integrate them into a unified metric—the \textbf{FED-Score}—by computing their product:
\begin{equation}
    \text{FED-Score} = \mathcal{S}_{\text{fid}} \times \mathcal{S}_{\text{align}} \times \mathcal{S}_{\text{reg}}.
\end{equation}

This multiplicative formulation ensures that a near-zero value in any single dimension is sufficient to suppress the entire FED-Score, so that all three dimensions must be simultaneously satisfied to achieve a high overall score.
We empirically validate that the FED-Score aligns significantly better with human perceptual preferences than conventional metrics in \cref{sec:human_correlation}.

\section{Experiments}
\label{sec:experiments}

\subsection{Experimental Setup}
\label{subsec:setup}

\noindent\textbf{Human Study Protocol.}
To validate metric reliability, we conduct a human correlation study using a Two-Alternative Forced Choice (2AFC) protocol. We randomly sample 2,760 image pairs, where each pair consists of editing results from two different models given the same source image and instruction. Three independent annotators are asked to select the better image within each pair from three perspectives: identity preservation, expression change magnitude, and overall quality. The final human preference is determined by majority voting, and we report the accuracy between each metric's algorithmic preference and the human consensus.

\noindent\textbf{Evaluated Models.}
To comprehensively assess the current landscape of facial expression editing, we benchmark 18 representative image editing models. All models are evaluated using their officially recommended default inference configurations to ensure a fair and reproducible comparison. Additionally, we evaluate our fine-tuned model, \textbf{Flux-Kontext-FED}, which is obtained by fine-tuning Flux-Kontext-dev on our proposed 20k+ training set, to validate the effectiveness of the curated training data.

\noindent\textbf{Evaluation Metrics.}
We evaluate all models using our proposed FED-Score and its constituent sub-metrics. To demonstrate the superiority of FED-Score over conventional approaches, we also compare against widely-used baseline metrics, including L2 distance, LPIPS\cite{zhang2018unreasonable}, CLIP-I, DINO Score\cite{liu2024grounding}, CLIP Score\cite{hessel-etal-2021-clipscore}, and ArcFace cosine similarity\cite{deng2019arcface}.

\noindent\textbf{Evaluation Protocol.}
All experiments are conducted on the full FED-Bench test set without subset splitting. Each model is evaluated under both \textit{simple instructions} (\eg, ``change the expression from neutral to happy'') and \textit{dense instructions} that describe fine-grained facial muscle movements, enabling a thorough assessment of instruction-following capabilities at different granularities.

\noindent\textbf{Implementation Details.}
For the MLLM-based scoring components (PQ, SC, and GTA), we uniformly adopt Gemini-2.5-Pro as the evaluation backbone to ensure consistency across all perceptual assessments. Identity Preservation is computed using ArcFace-R100 for face embedding extraction. The facial region mask $M_{\text{bg}}$ required for Background Consistency is obtained via Grounded SAM 2\cite{ravi2024sam2segmentimages,liu2024grounding}, which provides precise segmentation of the face region. For the REG metric, we employ LPIPS with a VGG-16 backbone, where the facial crop is directly obtained from the bounding box predicted by Grounding DINO.

\subsection{Human Alignment Test}
\label{sec:human_correlation}

\begin{table}[t]
\caption{Human alignment test. We report the accuracy between each metric's preference and the human consensus under a 2AFC protocol. $_\text{F}$: computed on the face region; $_\text{GT}$: compared with the ground truth image.}
\label{tab:human_consistency}
\centering
\scriptsize
\begin{subtable}[t]{0.22\textwidth}
\centering
\caption{ID Fidelity}
\label{tab:human_id}
\begin{tabular}{@{}lc@{}}
\toprule
Metrics & Acc. \\
\midrule
L2$_\text{F}$       & 0.6446 \\
DINO$_\text{F}$     & 0.6596 \\
LPIPS$_\text{F}$    & 0.6528 \\
CLIP-I$_\text{F}$   & 0.6384 \\
CLIP-T$_\text{F}$   & 0.5420 \\
\midrule
ArcFace$_\text{F}$  & \textbf{0.6606} \\
\bottomrule
\end{tabular}
\end{subtable}
\hfill
\begin{subtable}[t]{0.25\textwidth}
\centering
\caption{REG}
\label{tab:human_reg}
\begin{tabular}{@{}lc@{}}
\toprule
Metrics & Acc. \\
\midrule
L2$_\text{F,GT}$      & 0.6047 \\
DINO$_\text{F,GT}$    & 0.6818 \\
LPIPS$_\text{F,GT}$   & 0.7045 \\
CLIP-I$_\text{F,GT}$  & 0.5909 \\
ArcFace$_\text{F,GT}$ & 0.5455 \\
RPM-Modify$_\text{F}$   & 0.6154 \\
\midrule
LPIPS-Ratio$_\text{F}$ & \textbf{0.7386} \\
\bottomrule
\end{tabular}
\end{subtable}
\hfill
\begin{subtable}[t]{0.27\textwidth}
\centering
\caption{Overall Score}
\label{tab:human_overall}
\begin{tabular}{@{}lc@{}}
\toprule
Metrics & Acc. \\
\midrule
L2                    & 0.5711 \\
DINO$_\text{GT}$      & 0.6216 \\
DINO                  & 0.5816 \\
LPIPS$_\text{GT}$     & 0.6099 \\
LPIPS                 & 0.5384 \\
CLIP-I$_\text{GT}$    & 0.6850 \\
CLIP-I                & 0.6325 \\
CLIP-T                & 0.4480 \\
\midrule
FED-Score             & \textbf{0.7700} \\
\bottomrule
\end{tabular}
\end{subtable}
\hfill
\begin{subtable}[t]{0.22\textwidth}
\centering
\caption{FED-Score Abl.}
\label{tab:human_ablation}
\begin{tabular}{@{}lc@{}}
\toprule
Metrics & Acc. \\
\midrule
w/o Rule            & 0.7422 \\
w/o REG             & 0.7577 \\
w/o Fidelity        & 0.7379 \\
w/o Alignment       & 0.7279 \\
w/o Model           & 0.7202 \\
\midrule
FED-Score           & \textbf{0.7700} \\
\bottomrule
\end{tabular}
\end{subtable}
\end{table}

\noindent\textbf{Identity Preservation.}
As shown in \cref{tab:human_id}, ArcFace$_\text{F}$ achieves the highest agreement with human judgment among all face-region baselines, followed closely by DINO$_\text{F}$ and LPIPS$_\text{F}$. While all metrics surpass the random baseline of 0.5, their moderate absolute values (0.54--0.66) indicate that assessing identity preservation from face crops alone remains inherently challenging. Notably, CLIP-T$_\text{F}$ barely exceeds chance level, confirming that text-image similarity metrics are fundamentally unsuitable for identity evaluation. These results validate our choice of ArcFace cosine similarity as the ID component in FED-Score.

\noindent\textbf{Relative Expression Gain.}
As shown in \cref{tab:human_reg}, our proposed LPIPS-Ratio$_\text{F}$ achieves the highest accuracy (0.7386) for evaluating expression change magnitude, outperforming all GT-based baselines. Perceptual similarity measures (LPIPS$_\text{F,GT}$, DINO$_\text{F,GT}$) perform reasonably well, as they are inherently sensitive to fine-grained visual differences. RPM-Modify$_\text{F}$\cite{li2025balancingpreservationmodificationregion}, another ratio-based metric, achieves moderate accuracy (0.6154), falling between the perceptual baselines and identity-oriented metrics. In contrast, identity-oriented metrics like ArcFace$_\text{F,GT}$ and CLIP-I$_\text{F,GT}$ drop near random chance, suggesting that their feature spaces are not designed to capture expression change magnitude. The superiority of LPIPS-Ratio$_\text
{F}$ stems from its ratio-based formulation: by normalizing the actual 
expression change against the expected GT change, it directly quantifies 
relative editing effort and naturally penalizes both lazy editing (ratio $\ll$ 
1) and overfit editing (ratio $\gg$ 1).

\noindent\textbf{Overall Score.}
When evaluating holistic editing quality (\cref{tab:human_overall}), FED-Score achieves 0.77 agreement with human consensus, substantially outperforming the best single baseline CLIP-I$_\text{GT}$. Across all baselines, GT-based variants consistently surpass their source-target counterparts, reinforcing the value of reference-guided evaluation. Strikingly, CLIP-T falls below random chance, indicating that standalone text-image similarity is fundamentally unreliable for this task. These results confirm that the multi-dimensional design of FED-Score captures complementary evaluation aspects that no single metric can address alone.

\noindent\textbf{Ablation Study.}
We conduct an ablation study on FED-Score (\cref{tab:human_ablation}). Removing any single dimension consistently degrades performance, confirming that all components are essential. The model-based MLLM metrics prove most critical: removing them (w/o Model) causes the largest drop, as perceptual quality and semantic consistency cannot be captured by rule-based computation alone. Removing alignment and fidelity also leads to substantial degradation. Rule-based metrics likewise complement the MLLM components, showing that objective measurements of identity and background preservation remain valuable even alongside powerful vision-language models. REG exhibits the smallest individual impact, which is expected since it specifically targets expression change magnitude---a dimension less relevant in identity-focused and quality-focused evaluation pairs. Overall, this ablation validates that the synergy between rule-based computation and model-based perception is the key to FED-Score's superior human alignment.

\subsection{Benchmarking SOTA Models}
\label{subsec:benchtest}

\begin{table*}[t]
\caption{Benchmarking results on FED-Bench. Left: Dense instructions (ranked by Dense FED-Score). Right: Simple instructions (ranked by Simple FED-Score). We report raw mean values for sub-metrics and normalized FED-Score. BG$\downarrow$: lower is better. REG: optimal at 1.0. \textbf{Bold}: best; \underline{underline}: second best.}
\label{tab:benchmark}
\centering
\resizebox{\textwidth}{!}{%
\begin{tabular}{lccccccc|lccccccc}
\toprule
\multicolumn{8}{c|}{\textbf{Dense Instructions}} & \multicolumn{8}{c}{\textbf{Simple Instructions}} \\
\cmidrule(lr){1-8} \cmidrule(lr){9-16}
Method & ID & BG & PQ & SC & GTA & REG & Score & Method & ID & BG & PQ & SC & GTA & REG & Score \\
\midrule
Qwen-image-edit-plus        & .58 & 17.5 & 9.7 & 8.8 & \textbf{5.7} & 1.18 & \textbf{.469} & Qwen-image-edit-plus        & .63 & 16.8 & \underline{9.8} & 8.8 & \textbf{5.8} & 1.13 & \textbf{.492} \\
SeedDream 4.0\cite{seedream2025seedream40nextgenerationmultimodal}     & .62 & 15.5 & 9.5 & \underline{9.1} & \underline{4.3} & 1.37 & \underline{.379} & SeedDream 4.0     & \underline{.69} & 16.4 & 9.7 & 8.0 & 5.0 & 1.26 & \underline{.413} \\
FLUX.2 Pro\cite{flux-2-2025}        & .58 & 13.9 & \underline{9.8} & 9.0 & 4.0 & 1.37 & .377 & FLUX.2 Pro        & .58 & 14.0 & 9.6 & 8.8 & \underline{5.2} & 1.37 & .400 \\
Qwen-image-edit         & .45 & 19.6 & 9.3 & 8.9 & 4.2 & 1.37 & .337 & Qwen-image-edit-2511    & .44 & 15.3 & \underline{9.8} & \textbf{9.4} & 4.0 & 1.42 & .361 \\
FLUX-Kontext-FED  & .68 & \underline{7.3}  & 9.7 & 6.2 & 3.4 & \underline{0.95} & .332 & Qwen-image-edit         & .43 & 19.9 & 9.6 & 8.7 & 4.4 & 1.37 & .343 \\
FLUX-Kontext-Pro  & .52 & \underline{7.3}  & \underline{9.8} & 7.6 & 3.5 & \textbf{0.99} & .327 & Step1X v1p2       & .52 & 17.8 & 9.7 & \underline{9.3} & 3.1 & 1.41 & .333 \\
FLUX-Kontext-Max  & .50 & 9.7  & 9.7 & 7.7 & 3.5 & 1.07 & .320 & FLUX-Kontext-FED  & .68 & 7.8  & 9.7 & 6.5 & 2.9 & \underline{0.96} & .325 \\
Qwen-image-edit-2511\cite{wu2025qwenimagetechnicalreport}    & .46 & 15.8 & 9.5 & \textbf{9.3} & 3.4 & 1.43 & .317 & FLUX-Kontext-Max  & .49 & 8.1  & \textbf{9.9} & 5.2 & 3.7 & \textbf{1.03} & .259 \\
Step1X v1p2       & .56 & 17.1 & 9.4 & 7.7 & 3.0 & 1.31 & .303 & SeedEdit 3.0      & .48 & 12.8 & 9.1 & 7.6 & 2.7 & 1.56 & .239 \\
FLUX-Kontext-Dev\cite{labs2025flux1kontextflowmatching}  & \underline{.72} & 23.8 & 9.6 & 6.7 & 3.0 & 0.67 & .243 & FLUX-Kontext-Pro  & .55 & 8.5  & \textbf{9.9} & 4.8 & 3.4 & \textbf{0.97} & .227 \\
SeedEdit 3.0\cite{wang2025seededit30fasthighquality}      & .49 & 11.6 & 7.8 & 7.4 & 1.9 & 1.55 & .203 & Bagel             & .53 & 13.1 & 5.4 & 4.1 & 2.4 & 1.29 & .163 \\
UniWorld-v2\cite{li2025uniworldv2reinforceimageediting}          & .37 & 31.6 & 9.1 & 7.6 & 2.5 & 1.45 & .201 & Step1X            & .32 & 17.3 & 9.7 & 5.6 & 2.3 & 1.67 & .149 \\
DreamOmni2\cite{xia2025dreamomni2}         & \textbf{.81} & 24.1 & 9.6 & 5.0 & 2.6 & 0.45 & .168 & OmniGen2           & .44 & 77.6 & 8.3 & 4.1 & 3.6 & 1.61 & .122 \\
OmniGen2\cite{wu2025omnigen2}           & .56 & 63.7 & 7.3 & 6.3 & 2.6 & 1.52 & .155 & FLUX-Kontext-Dev  & \textbf{.86} & \textbf{4.8}  & \underline{9.8} & 2.3 & 3.1 & 0.35 & .120 \\
FLUX-Kontext-Fill & .11 & \textbf{5.1}  & \textbf{9.9} & 5.5 & 1.7 & 1.56 & .155 & UniWorld-v2          & .30 & 34.9 & 8.6 & 3.4 & 2.7 & 1.60 & .110 \\
Step1X\cite{liu2025step1x}            & .33 & 16.3 & 7.3 & 6.9 & 1.3 & 1.72 & .127 & FLUX-Kontext-Fill & .11 & \underline{5.1}  & 9.7 & 3.3 & 1.1 & 1.52 & .096 \\
Bagel\cite{deng2025bagel}             & .42 & 47.2 & 6.5 & 6.8 & 1.8 & 1.57 & .115 & DreamOmni2         & .45 & 36.9 & 8.9 & 1.2 & 2.0 & 1.40 & .049 \\
InstructPix2Pix\cite{brooks2023instructpix2pix}   & .08 & 47.0 & 0.0 & 0.0 & 0.0 & 2.56 & .001 & InstructPix2Pix   & .08 & 41.2 & 0.2 & 0.1 & 0.1 & 2.42 & .004 \\
\bottomrule
\end{tabular}%
}
\vspace{-16pt}
\end{table*}

\noindent\textbf{Overall Rankings.}
\cref{tab:benchmark} presents the comprehensive benchmarking results across 18 models. Qwen-Image-Edit-Plus consistently ranks first under both Dense and Simple instructions, achieving the highest GTA by a significant margin while maintaining strong performance across all other dimensions. SeedDream 4.0 and FLUX.2 Pro secure the second and third positions in both settings, with SeedDream 4.0 exhibiting particularly strong ID preservation and FLUX.2 Pro delivering well-balanced sub-metric scores. Notably, the top-3 ranking remains stable across instruction types, suggesting that these models possess genuine comprehensive editing proficiency rather than excelling at a particular instruction granularity. It is also worth highlighting that FLUX-Kontext-FED, fine-tuned on our proposed training set, ranks 5th under Dense instructions and 7th under Simple instructions, substantially outperforming the original FLUX-Kontext-Dev, which validates the practical value of our released training data.

\noindent\textbf{Fidelity vs.\ Alignment Trade-off.}
A key insight revealed by FED-Score is that no single model excels across all dimensions simultaneously. FLUX-Kontext-Dev and DreamOmni2 achieve the highest ID scores, yet their extremely low REG exposes a classic \textit{lazy editing} pattern: preserving identity by barely modifying the expression. On the opposite extreme, InstructPix2Pix exhibits severe \textit{overfit editing} with REG$>$2.5 while catastrophically degrading visual quality (PQ$\approx$0). FLUX-Kontext-Fill presents yet another failure mode---best BG and PQ but worst ID, generating high-quality images that completely lose the subject's identity. Under conventional single-metric evaluation, lazy editors would be incorrectly ranked at the top based on ID or BG alone. FED-Score effectively penalizes such dimensional imbalances, ensuring that only models with genuinely balanced performance receive high overall scores.

\noindent\textbf{Dense vs.\ Simple Instructions.}
The ranking shifts between the two settings expose notable differences in instruction-following capability. FLUX-Kontext-Pro drops significantly from Dense to Simple, with its SC nearly halving, suggesting it benefits disproportionately from the richer guidance in dense instructions. Conversely, FLUX-Kontext-Dev's lazy editing tendency intensifies under simple instructions, as its REG further deteriorates. The Alignment metrics (SC, GTA) exhibit the largest cross-setting variance, confirming that fine-grained instruction following remains the primary bottleneck for current models.

\begin{figure}[t]
    \centering
    \includegraphics[width=\linewidth]{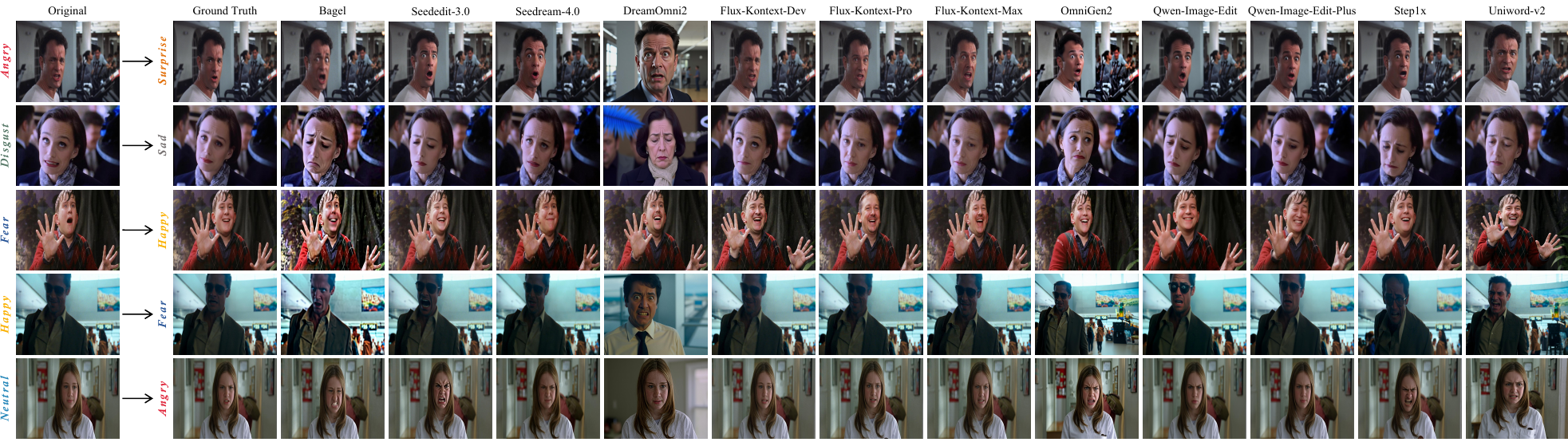}
    \caption{The qualitative comparison of facial expression editing on our FED-Bench. Each row shows a different editing task with the original image, ground truth target, and results from part of evaluated models. From left to right: original image, ground truth, followed by the evaluation results.}
    \label{fig:case_study}
    \vspace{-8pt}
\end{figure}

\noindent\textbf{Qualitative Results.} \cref{fig:case_study} presents qualitative comparisons that corroborate the quantitative findings. The FLUX-Kontext series produces high-fidelity outputs with well-preserved backgrounds and identity, yet their expression transformations remain subtle---a visual manifestation of the \textit{lazy editing} pattern identified in our quantitative analysis. In contrast, Qwen-Image-Edit-Plus generates clear and natural target emotions closely aligned with the ground truth, consistent with its top-ranked FED-Score. Models such as DreamOmni\cite{xia2025dreamomni2} and OmniGen2\cite{wu2025omnigen2} introduce significant non-expression modifications including background alterations and lighting changes, visually confirming the fidelity-alignment trade-off revealed by the benchmarking results. 
Furthermore, based on qualitative analysis and REG experimental data, most models exhibit overfit editing.
The detailed qualitative cases and related analysis results are presented in the supplementary materials.
\section{Conclusion}
\label{sec:conclusion}

In this paper, we present FED-Bench, a comprehensive benchmark for facial expression image editing. We construct 747 high-quality triplets via a scalable multi-stage pipeline, and propose FED-Score, a cross-granularity protocol that disentangles evaluation into Fidelity, Alignment, and Relative Expression Gain, effectively penalizing lazy editing and overfit editing. Human alignment experiments confirm that FED-Score substantially outperforms all traditional metrics. Benchmarking 18 image editing models and revealing that balancing fidelity with accurate expression manipulation remains challenging, with fine-grained instruction following as the primary bottleneck. We additionally release a 20k+ in-the-wild training set and validate its utility by fine-tuning FLUX-Kontext-Dev, yielding notable gains in both expression accuracy and fidelity.

\bibliographystyle{splncs04}
\bibliography{main}

\end{document}